\pdfoutput=1

\documentclass[11pt]{article}

\usepackage{graphicx}
\usepackage{multirow}

\usepackage{colortbl}
\usepackage{times}
\usepackage[T2A,T1]{fontenc}
\usepackage[utf8]{inputenc} 
\usepackage{microtype} 
\usepackage{inconsolata} 
\usepackage{newtxtext}
\usepackage{newtxmath}

\usepackage[polish, ukrainian, russian, german, dutch, english]{babel}

\usepackage{tabularx, booktabs} 
\usepackage{array} 
\usepackage{caption} 
\usepackage{subcaption} 
\usepackage{siunitx}
\usepackage{threeparttable}
\usepackage[table,xcdraw]{xcolor} 

\usepackage[final]{acl}

\definecolor{PosDiff}{HTML}{006400} 
\definecolor{NegDiff}{HTML}{8B0000} 

\definecolor{lower}{HTML}{ADD8E6} 
\definecolor{mid}{HTML}{FFFF99}   
\definecolor{upper}{HTML}{FFB6C1} 

\title{Attention on Multiword Expressions: A Multilingual Study of BERT-based Models with Regard to Idiomaticity and Microsyntax}

\author{Iuliia Zaitova, Vitalii Hirak, Badr M. Abdullah, \\ {\bf Dietrich Klakow}, {\bf Bernd Möbius}, {\bf Tania Avgustinova} \\
        Saarland University, Germany \\
        \texttt{izaitova@lsv.uni-saarland.de}
}

\renewcommand{\arraystretch}{0.9} 

\begin{document}
\maketitle
\begin{abstract}
This study analyzes the attention patterns of fine-tuned encoder-only models based on the BERT architecture (BERT-based models) towards two distinct types of Multiword Expressions (MWEs): idioms and microsyntactic units (MSUs). Idioms present challenges in semantic non-compositionality, whereas MSUs demonstrate unconventional syntactic behavior that does not conform to standard grammatical categorizations.
We aim to understand whether fine-tuning BERT-based models on specific tasks influences their attention to MWEs, and how this attention differs between semantic and syntactic tasks. We examine attention scores to MWEs in both pre-trained and fine-tuned BERT-based models. We utilize monolingual models and datasets in six Indo-European languages — English, German, Dutch, Polish, Russian, and Ukrainian.
Our results show that fine-tuning significantly influences how models allocate attention to MWEs. Specifically, models fine-tuned on semantic tasks tend to distribute attention to idiomatic expressions more evenly across layers. Models fine-tuned on syntactic tasks show an increase in attention to MSUs in the lower layers, corresponding with syntactic processing requirements. 
\end{abstract}
\section{Introduction}

\begin{figure}[t!]
\centering\includegraphics[width=1\linewidth]{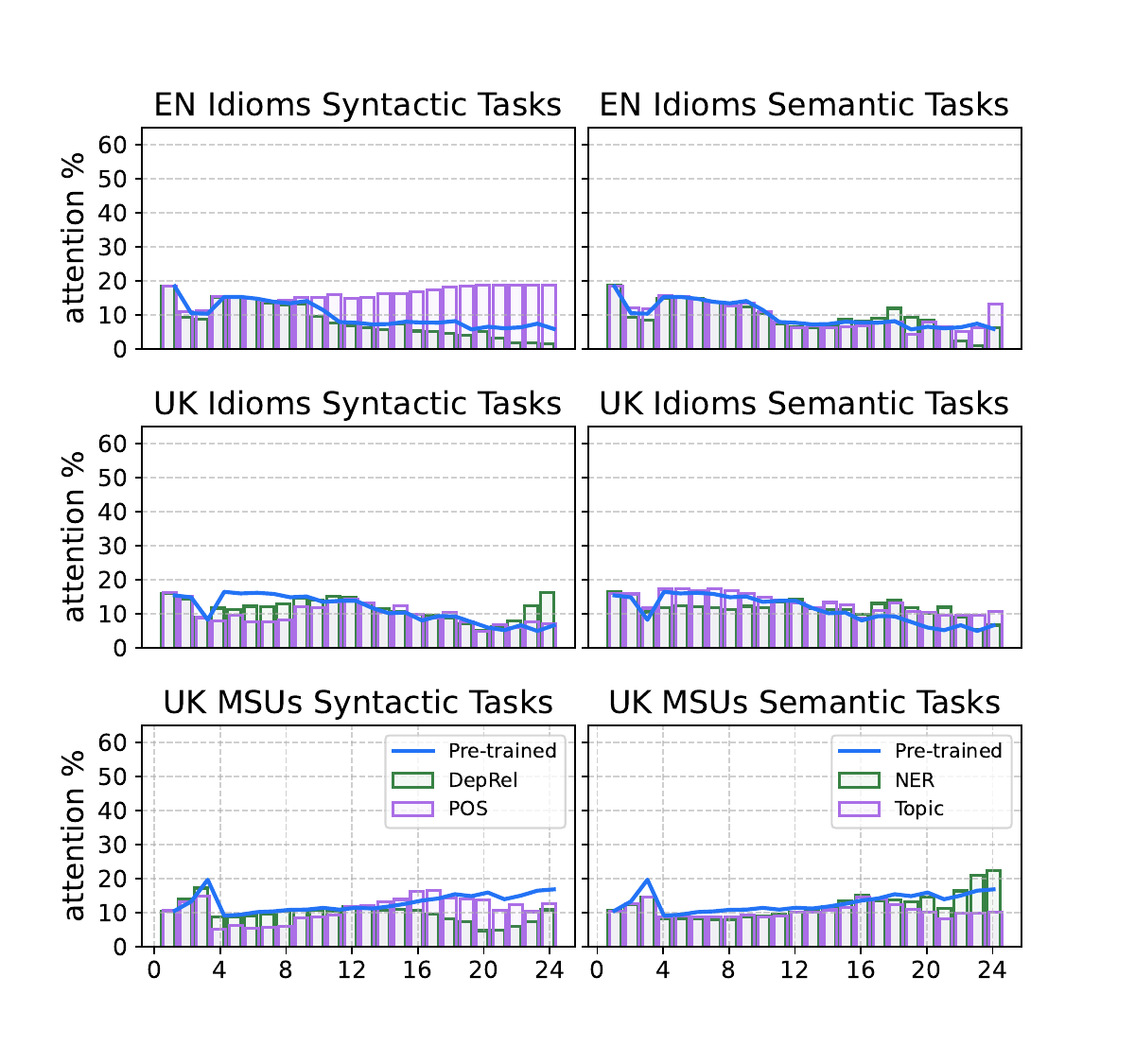}
     \caption{Layer-wise attention distribution in BERT-based models for idioms and microsyntactic units (MSUs). Results shown for English (EN) and Ukrainian (UK). Models fine-tuned on syntactic tasks (Dependency Relation Classification -- DepRel, Part-of-Speech Tagging -- POS) are on the left, and semantic tasks (Named Entity Recognition -- NER, Topic Classification -- Topic) are on the right. The y-axis shows the percentage of attention scores directed \emph{from other sentence tokens towards Multiword Expressions}, with higher percentages indicating stronger attention focus.} 
    \label{fig:attention_results}
\end{figure}

Attention mechanisms in Natural Language Processing (NLP) enhance the ability of models to focus on relevant aspects of input data by adjusting the model's focus dynamically. Such capabilities are foundational in models like BERT \cite{devlin-etal-2019-bert}, which utilize attention mechanisms to manage deep contextual understanding and complex linguistic phenomena.

Multiword Expressions (MWE) consist of two or more (lexical) components and function as a single unit. This is characteristic of idioms, collocations, and formulaic expressions, all of which exhibit a degree of semantic cohesion that distinguishes them from literal combinations of the contained words \cite{MultiWordExpressionsandMorphology, shwartz-dagan-2019-still}. \citet{warren_2005} notes that such expressions are pervasive in language use, often outnumbering single words. 
In this study, we are specifically interested in two distinct MWE types, each presenting its unique challenges: idioms and microsyntactic units.

Idioms, such as \textit{spill the beans} or 
\textit{das Handtuch werfen} ("throw in the towel") in German, embody the challenge of semantic non-compositionality. These expressions, characterized by meanings that cannot be deduced simply from their literal components, require a deeper understanding of language beyond its surface structure. The complexity of idioms poses significant challenges for NLP systems, as they must navigate these non-literal, often figurative expressions to achieve accurate understanding and generation of language \cite{baldwin2010multiword, fadaee2018examining, 10.1162/tacl_a_00442, tan-jiang-2021-bert}.

Microsyntactic units (MSU), such as \textit{all the same} or \selectlanguage{russian}\textit{тем не менее} \selectlanguage{english} (translit.: "tem ne menee", "nevertheless") in Russian, fall out of standard grammatical categorization and analysis, due to syntactically unpredictable behavior \cite{iomdin2015}. For instance, the MSU \textit{all the same} can function in multiple ways: it may express persistence despite a fact, as in 'She was kind, but all the same she terrified me', or indifference, as in \textit{It is all the same to me whether you stay or go} \cite{iomdin-2016-microsyntactic}. \citet{Avgustinova2019} highlight the role of MSUs in challenging the boundaries between lexicon and grammar, noting the need for advanced parsing strategies that can adapt to their unique syntactic structure.

In this paper, we analyze the attention patterns of both pre-trained and fine-tuned encoder-only models based on the
BERT architecture (BERT-based models) towards two distinct MWE types: idioms and MSUs. We adopt the approach of \citet{jang-etal-2024-study}, who examine whether the attention scores in BERT change during the fine-tuning process for downstream tasks. Extending their methodology to MWEs allows us to contrast how these models handle the semantic non-compositionality in idioms against the syntactic non-conformity in MSUs. To validate our results, we employ monolingual models and datasets in six Indo-European languages coming from two language groups: Slavic and Germanic. The languages are English (\textbf{EN}), German (\textbf{DE}), Dutch (\textbf{NL}), Polish (\textbf{PL}), Russian (\textbf{RU}), and Ukrainian (\textbf{UK}). The choice of these languages is motivated by extensive linguistic resources and pre-trained models available for them, which is essential for our analyses. While we acknowledge that the language sample is not typologically diverse and does not include languages outside of Indo-European language family, leveraging these specific languages allows for a more controlled analysis within the scope of our research.

Our results indicate that fine-tuning enhances the models' ability to appropriately focus attention on MWEs, with distinct patterns observed between idioms and MSUs. Models fine-tuned on semantic tasks show a more even distribution of attention to idiomatic expressions, reflecting the need for integrating information across layers. In contrast, models fine-tuned on syntactic tasks exhibit increased attention to MSUs in lower to middle layers, according to syntactic processing requirements. These findings suggest that attention mechanisms in BERT-based models adapt during fine-tuning to better handle different types of MWEs, aligning with the linguistic characteristics of idioms and MSUs.

To facilitate further research, we have made our datasets and fine-tuned models publicly available at \texttt{github.com/IuliiaZaitova/mwe-attention}.

\section{Related Work}
\label{related_work}

\textbf{Interpreting BERT's Attention.} While the attention mechanisms in BERT-based models have been extensively studied, some works stand out for their relevance to our current study.

\citet{tenney2019bertrediscoversclassicalnlp} investigate how linguistic information is represented across different layers within BERT, discovering that the model architecture implicitly mirrors the classical NLP pipeline, contradicting the often criticized "black box" nature of such models. Their findings indicate that lower layers of BERT are better at encoding local syntactic information, while higher layers progressively engage with more complex semantic processing.
In addition, they observe that syntactic information is more localizable, with weights related to syntactic tasks tending to be focused on a few layers, while information related to semantic tasks is generally spread across the entire network. These findings suggests that despite BERT's holistic training approach, it may maintain an interpretable and hierarchical structure.

\citet{jang-etal-2024-study} specifically examined how BERT's attention scores vary with lexical categories during the fine-tuning process on downstream tasks. Their study explores the model's behavior during the fine-tuning phase on the GLUE benchmark tasks \cite{wang-etal-2018-glue}, hypothesizing that BERT's attention mechanism is selectively sensitive to the lexical category of tokens — with increased attention to content words for semantic tasks and to function words for syntactic tasks. Their findings confirm that BERT's attention is not uniformly distributed but is instead strategically adjusted to emphasize relevant lexical categories based on the task, demonstrating a certain level of linguistic adaptability.

\textbf{MWE Processing in Language Models.} Prior studies have explored both the extent to which Language Models understand MWEs and their different types \cite{kurfali-2020-travis, walsh-etal-2022-berts, miletic-walde-2024-semantics, dankers-etal-2022-transformer, rambelli-etal-2023-frequent, tian-etal-2023-idioms}.  
Further research has investigated how these models can be fine-tuned to improve their performance on the classification of MWEs \cite{boisson-etal-2022-cardiffnlp, avram-etal-2023-romanian, bui-savary-2024-cross}.

A work that has a particular relevance is \citet{zaitova-etal-2023-microsyntactic}. This study proposes an approach to detect MSUs by using cosine similarity retrieved from five Word Embedding Models (WEMs), and evaluates how well these models capture syntactic idiosyncrasies. The results demonstrate the effectiveness of WEMs in capturing MSUs across six Slavic languages. Additionally, it shows that WEMs adapted for syntax-based tasks consistently outperform other WEMs at the task.

In spite of these contributions, there is still a gap in understanding how BERT-based models attend to different types of MWEs, particularly in a multilingual context. Previous studies often focus on a single language or do not consider different types of MWEs. Our study addresses these limitations by analyzing how fine-tuning affects attention to MWEs in BERT-based models across six languages. 

\section{Methodology}

\subsection{Datasets}

We conducted our experiments with BERT-based models using idiom and MSU datasets. While the idiom dataset includes two groups of Indo-European languages -- Germanic (DE, EN, NL) and Slavic (PL, RU, UK), the MSU dataset only includes the Slavic languages (PL, RU, UK).

An example of the data in all languages is given in Table ~\ref{tab:idioms_microsyntax}. To ensure a fair comparison between idioms and MSUs, we selected or created a subset of 227 idioms in context for each language where possible, matching the number of MSUs in the MSU dataset to allow for balanced analyses.

\begin{table*}[htbp]
\centering
\captionsetup{width=\textwidth}
\small 
\renewcommand{\arraystretch}{1.2}
\begin{tabularx}{\textwidth}{
    >{\raggedright\arraybackslash}p{0.05\textwidth}
    >{\raggedright\arraybackslash}p{0.08\textwidth}
    >{\raggedright\arraybackslash}X
    >{\raggedright\arraybackslash}X}
\toprule
\textbf{Lang} & \textbf{Type} & \textbf{Sentence} & \textbf{English Translation} \\
\midrule
\rowcolor{gray!10} EN & Idiom & They covered the whole field \textbf{from A to Z} in eight classes. & - \\
DE & Idiom & \textbf{\selectlanguage{german}Ihre große Liebe} sei Jonathon, sagt Sarah. & Sarah then says that Jonathon is her \textbf{great love}. \\
\rowcolor{gray!10} NL & Idiom & \textbf{\selectlanguage{dutch}Af en toe} verzorgde ze nog een gastoptreden. & \textbf{Now and then}, she would make a guest appearance. \\
 PL & Idiom & \textbf{Cały czas} było mi zimno z nim. & I was \textbf{constantly} cold because of him. \\
\rowcolor{gray!10} PL & MSU & \textbf{Z trudem} kojarzył i pojmował, co do niego mówią. & He \textbf{barely} understood what was said to him. \\
 UK & Idiom & \selectlanguage{ukrainian} Для нього робота — це \textbf{альфа і омега} всього життя. & For him, work is the \textbf{be-all and end-all} of life. \\
\rowcolor{gray!10} UK & MSU & \selectlanguage{ukrainian} Я \textbf{\selectlanguage{ukrainian}все ще} сподівався на банальну аварію. & I was \textbf{still} hoping it was just a mundane accident. \\
RU & Idiom & \selectlanguage{russian} За что позор за позором  \textbf{\selectlanguage{russian}валится на мою голову}? & Why does disgrace after disgrace \textbf{fall on my head}? \\
\rowcolor{gray!10} RU & MSU & \selectlanguage{russian} Я \textbf{\selectlanguage{russian}всё время} думал о тебе, день и ночь. & I thought about you \textbf{all the time}, day and night. \\
\bottomrule
\end{tabularx}
\caption{Idioms and microsyntactic units in context.}
\label{tab:idioms_microsyntax}
\end{table*}

\subsubsection{Microsyntactic Unit Dataset}

The Slavic MSU dataset \cite{zaitova-etal-2023-microsyntactic}\footnote{\url{https://huggingface.co/datasets/izaitova/slavic_fixed_expressions}} was compiled using the list of MSUs provided in the Russian National Corpus\cite{ruscorpora}. The selection process focused on the most frequently occurring MSUs, resulting in a total of 227 instances for Russian. 

These 227 Russian MSUs are accompanied by their translational equivalents and parallel bilingual context sentences across five Slavic languages. The translational equivalents were manually sourced from the parallel sub-corpora of the Russian National Corpus and the Czech National Corpus \cite{machalek2020kontext}, generating parallel sets for the analysis. 
In our study, we only use the MSU sets for Polish, Russian, and Ukrainian.

\subsubsection{Idiom Dataset}

\paragraph{English, German, Dutch, and Polish Idiom Dataset.}

For the languages EN, DE, NL, and PL, we used the ID10M dataset \cite{tedeschi-etal-2022-id10m}, which provides idiom annotations in 10 languages. The dataset was developed as part of a complete framework for idiom identification in several languages. It includes automatically created training and development data with idioms, their context sentences, and their annotations. Additionally, the test sets for four languages were curated manually. Among the languages in our analysis, the test sets were available for EN and DE.

\paragraph{Russian Idiom Dataset.}

For Russian, we used 85 idioms from the dataset by \citet{Aharodnik2018DesigningAR}. The remaining 142 idioms in context were manually retrieved from the "Academic Dictionary of Russian Phraseology" \cite{baranov2015}, ensuring comprehensive coverage. The resulting idioms were proofread by two native speakers (a 28 year old female, and a 24 year old male), who are also professional linguists.

\paragraph{Ukrainian Idiom Dataset.}

For Ukrainian, due to the absence of pre-existing idiom datasets, we created our own dataset by generating idioms with OpenAI's ChatGPT, specifically using the GPT-4 model. Each idiom was subsequently verified by a native Ukrainian speaker (age: 24; gender: male), who is also a professional linguist to ensure its accuracy.

\subsection{Models}

We used six large 24-layer encoder-only transformer models based on the BERT architecture trained using only the masked language modeling objective on monolingual texts (BERT-based models). Specifically, for English we utilized BERT-large-cased, for German -- GBERT-large \cite{chan-etal-2020-germans}, for Dutch -- RobBERT-large \cite{delobelle2023robbert2023conversion}, for Polish -- HerBERT-large-cased \cite{mroczkowski-etal-2021-herbert}, for Russian -- ruBERT-large \cite{zmitrovich2023family}, and for Ukrainian -- Liberta-large \cite{haltiuk-smywinski-pohl-2024-liberta}.

\subsection{Fine-tuning Process}

To analyze how task-specific training affects the attention mechanisms of pre-trained models when processing idioms and MSUs, we fine-tuned each model on two syntactic and two semantic NLP tasks. The selection of tasks was guided not only by the availability of fine-tuning datasets in the studied languages, but also by their relevance to linguistic properties of MWEs.

\subsubsection{Syntactic Tasks}

\textbf{Dependency Relation Classification (DepRel)} predicts the dependency relation tag for each token in a sentence. This task assesses the model's ability to understand syntactic relationships between tokens, which is crucial for parsing the often non-standard structures of MWEs. 
For DepRel, we used datasets based on Universal Dependencies \cite{nivre-etal-2020-universal}.

\textbf{Part-of-Speech (POS) Tagging} assigns grammatical categories to each token, testing the model's grasp of syntactic roles and morphological forms. POS tagging is essential for handling tokens that may have unconventional syntactic functions in idiomatic versus literal contexts. Datasets for this task are also sourced from the Universal Dependencies. 

\subsubsection{Semantic Tasks}

 \textbf{Named Entity Recognition (NER)} identifies and classifies entities in texts into categories like people, organizations, and locations, evaluating the model's ability to extract semantic information. The task of NER is relevant to our study goal because MWEs are often culture-specific and can include named entities. 
 We utilized the WikiANN dataset \cite{pan-etal-2017-cross} for this task. 
 
\textbf{Topic Classification (Topic)} assigns sentences to predefined topics based on content, requiring understanding of broader semantic context. Topic classification is relevant since MWEs often require broader context to be interpreted correctly. The SIB-200 dataset \cite{adelani2023sib200} was used for this task. 

\subsubsection{Training and Evaluation}

The datasets were divided into training, development, and test sets, with sizes varying depending on the language and dataset. Table \ref{tab:stats} shows the number of training samples used for fine-tuning the models on each downstream task across the six languages studied. 

All models were fine-tuned for 10 epochs using the Hugging Face Transformers library \cite{wolf-etal-2020-transformers}, with the development set for validation and hyperparameter tuning. We evaluated the models on the test set using the F1 score to ensure consistent comparison across tasks. The models achieved competitive performance, with F1 scores exceeding 0.75 across all tasks and languages.

\begin{table}[h!] 

\centering 
\begin{tabular}{lcccc}
\toprule \textbf{Lang} & \textbf{DepRel} & \textbf{POS} & \textbf{NER} & \textbf{Topic} \\ \midrule 
\rowcolor{gray!10} EN & 5000 & 7000 & 5000 & 701 \\
DE & 5000 & 7000 & 5000 & 701 \\
\rowcolor{gray!10} NL & 5000 & 7000 & 5000 & 701 \\
PL & 5000 & 7000 & 5000 & 701 \\
\rowcolor{gray!10} RU & 5000 & 5400 & 7000 & 701 \\
UK & 5496 & 5000 & 7000 & 701 \\
 \bottomrule 
\end{tabular}
\caption{Number of training samples for fine-tuning by task.} 
\label{tab:stats} 
\end{table}

\section{Experimental Setup}

\subsection{Data Preprocessing and Setup}
Each context sentence with an MWE is tokenized using the pre-trained model tokenizer (e.g., BERT-large-cased tokenizer for English). This ensures consistency and accuracy in mapping MWE tokens to attention vectors. After tokenization, the inputs are fed into both the pre-trained and the fine-tuned models, and the attention outputs are extracted for analysis.

To maintain alignment between the tokens of the MWEs and the attention weights, we carefully handle cases where MWEs are split into subword tokens. We aggregate the attention weights corresponding to all subword tokens that compose an MWE, treating them as a single unit in our analysis.

\subsection{Attention Extraction}
For each model, we extract the multi-head attention matrices from all layers during the forward pass. These matrices represent the attention weights that each token in the input sequence assigns to every other token, including itself. Specifically, for a model with $L$ layers and $H$ attention heads per layer, we obtain $L \times H$ attention matrices per input sequence.

To simplify our analysis, we average the attention matrices across the $H$ heads in each layer. This results in a single averaged attention matrix per layer of size $T \times T$, where $T$ is the length of the tokenized input sequence. By examining these averaged matrices, we can analyze how attention is distributed across the model's layers.

\subsection{Attention Analysis}
To quantify the attention patterns related to MWEs, we adopt and extend the metrics proposed by \citet{jang-etal-2024-study}. Specifically, we additionally study the layer-wise changes in attention distribution across MWE categories both within MWE and from context to MWE.

Our analysis focuses on the following aspects:

\begin{itemize}
    \item \textbf{Attention from Context to MWEs}: For each layer in the model, we compute the average attention scores directed from all other tokens in the sentence towards tokens of MWEs. 
    \item \textbf{Attention within MWEs}: We also analyze the attention among tokens within MWE, calculating the average attention that MWE tokens pay to one another at each layer.
    \item \textbf{Impact of Fine-tuning}: We compare the pre-trained models to fine-tuned models and assess how fine-tuning on syntactic or semantic tasks changes the model's attention to MWEs. Moreover, we analyze the direction of changes in attention scores (positive or negative) for all models and both categories of MWEs.
        \item \textbf{Attention Differences between types of MWEs}: We look at the attention distributions for idioms and MSUs to compare the patterns for two types of MWEs. 
    \item \textbf{Language-Specific Attention Patterns}: We examine how different language-specific BERT-based models process MWEs to identify similarities and differences in attention patterns across languages and model architectures.
\end{itemize}

\section{Results and Discussion}

In this section, we present our analysis of the attention patterns by the pre-trained and fine-tuned BERT-based models when processing two types of MWEs: idioms and MSUs across different layers.

\begin{figure}[t!]
\centering\includegraphics[width=1\linewidth]{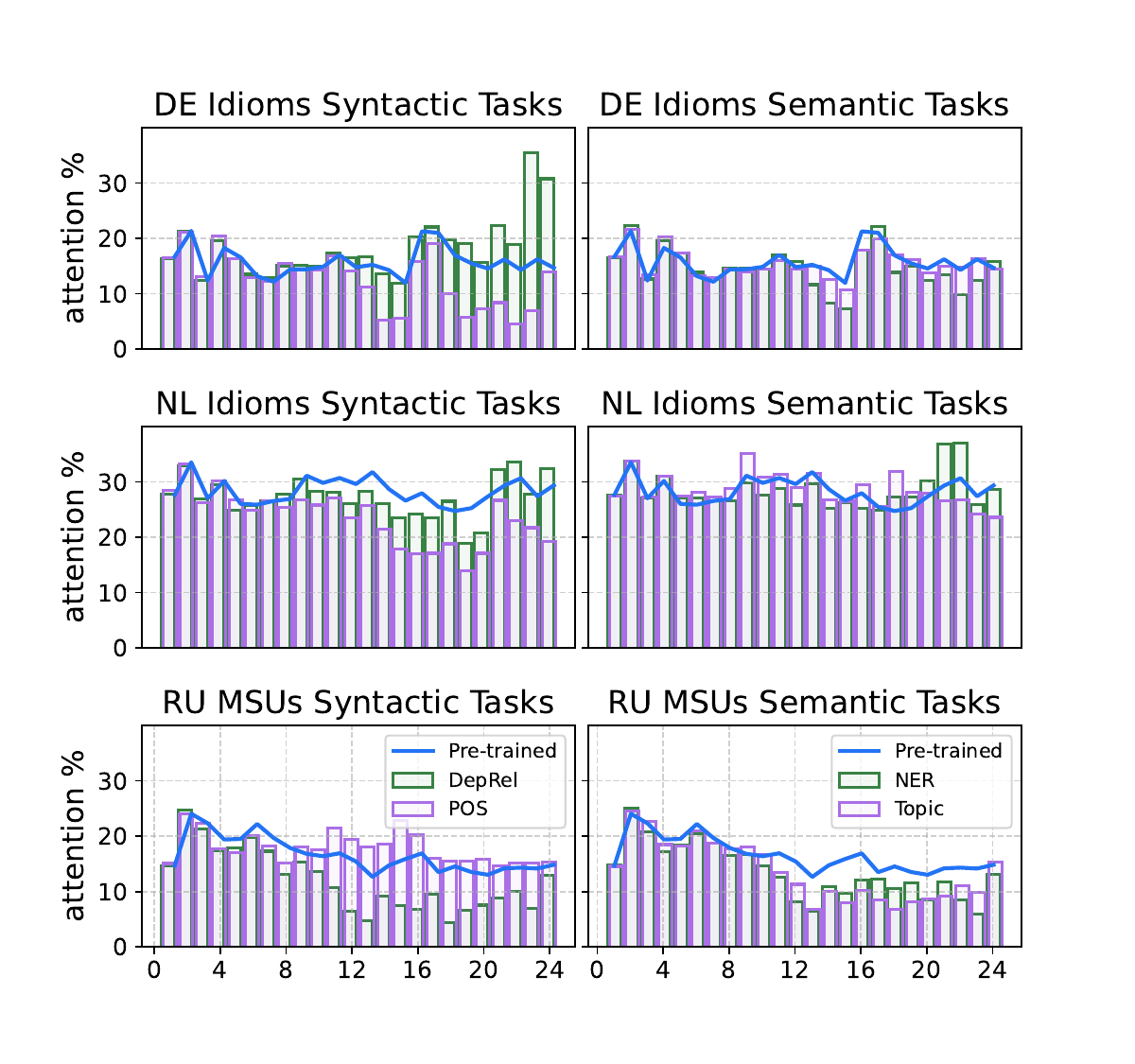}
     \caption{Layer-wise attention distribution within Multiword Expressions (MWEs) in German (DE), Dutch (NL), and Russian (RU) BERT-based models. The graphs show attention distribution within tokens of idioms and microsyntactic units (MSUs), comparing pre-trained models with those fine-tuned on syntactic tasks (Dependency Relation Classification -- DepRel, Part-of-Speech Tagging -- POS) and semantic tasks (Named Entity Recognition -- NER, Topic Classification -- Topic). The y-axis represents the percentage of attention between tokens within the same MWE.} 
    \label{fig:self_attention_plots}
\end{figure}

\begin{table*}
\begin{subtable}{1\textwidth}
\sisetup{table-format=1.2} 
\centering
\resizebox{\textwidth}{!}{%
\begin{tabular}{@{}l*{18}{c}@{}}
\toprule
\textbf{Task} & \multicolumn{3}{c}{\textbf{EN}} & \multicolumn{3}{c}{\textbf{DE}} & \multicolumn{3}{c}{\textbf{NL}} & \multicolumn{3}{c}{\textbf{PL}} & \multicolumn{3}{c}{\textbf{RU}} & \multicolumn{3}{c}{\textbf{UK}} \\
\cmidrule(lr){2-4}\cmidrule(lr){5-7}\cmidrule(lr){8-10}\cmidrule(lr){11-13}\cmidrule(lr){14-16}\cmidrule(lr){17-19}
 & T1 & T2 & T3 & T1 & T2 & T3 & T1 & T2 & T3 & T1 & T2 & T3 & T1 & T2 & T3 & T1 & T2 & T3 \\
\midrule
Pre-trained & \cellcolor{lower}{1} & \cellcolor{lower}{4} & \cellcolor{lower}{5} & \cellcolor{lower}{3} & \cellcolor{lower}{2} & \cellcolor{lower}{1} & \cellcolor{mid}{\textbf{12}} & \cellcolor{lower}{3} & \cellcolor{lower}{7} & \cellcolor{lower}{3} & \cellcolor{lower}{2} & \cellcolor{lower}{5} & \cellcolor{lower}{1} & \cellcolor{lower}{2} & \cellcolor{mid}{\textbf{10}} & \cellcolor{mid}{\textbf{4}} & \cellcolor{lower}{6} & \cellcolor{lower}{5}\\
\midrule
DepRel & \cellcolor{lower}{1} & \cellcolor{lower}{5} & \cellcolor{lower}{4} & \cellcolor{lower}{3} & \cellcolor{lower}{2} & \cellcolor{lower}{1} & \cellcolor{lower}{5} & \cellcolor{lower}{3} & \cellcolor{lower}{6} & \cellcolor{lower}{3} & \cellcolor{lower}{2} & \cellcolor{lower}{5} & \cellcolor{lower}{1} & \cellcolor{lower}{3} & \cellcolor{lower}{2} & \cellcolor{lower}{1} & \cellcolor{mid}{\textbf{11}} & \cellcolor{mid}{\textbf{12}}\\
POS & \cellcolor{lower}{1} & \cellcolor{mid}{\textbf{11}} & \cellcolor{lower}{4} & \cellcolor{lower}{3} & \cellcolor{lower}{2} & \cellcolor{lower}{5} & \cellcolor{mid}{\textbf{12}} & \cellcolor{lower}{7} & \cellcolor{lower}{3} & \cellcolor{lower}{3} & \cellcolor{lower}{2} & \cellcolor{lower}{1} & \cellcolor{lower}{1} & \cellcolor{lower}{2} & \cellcolor{mid}{\textbf{9}} & \cellcolor{lower}{1} & \cellcolor{lower}{2} & \cellcolor{mid}{\textbf{12}}\\
\midrule
NER & \cellcolor{lower}{1} & \cellcolor{lower}{5} & \cellcolor{lower}{6} & \cellcolor{lower}{3} & \cellcolor{lower}{6} & \cellcolor{lower}{2} & \cellcolor{mid}{\textbf{12}} & \cellcolor{lower}{7} & \cellcolor{lower}{6} & \cellcolor{lower}{3} & \cellcolor{lower}{2} & \cellcolor{lower}{5} & \cellcolor{lower}{1} & \cellcolor{lower}{2} & \cellcolor{lower}{3} & \cellcolor{lower}{1} & \cellcolor{lower}{2} & \cellcolor{mid}{\textbf{12}}\\
Topic & \cellcolor{lower}{1} & \cellcolor{lower}{4} & \cellcolor{lower}{5} & \cellcolor{lower}{3} & \cellcolor{lower}{1} & \cellcolor{lower}{2} & \cellcolor{lower}{3} & \cellcolor{lower}{7} & \cellcolor{lower}{5} & \cellcolor{lower}{3} & \cellcolor{lower}{2} & \cellcolor{lower}{5} & \cellcolor{lower}{1} & \cellcolor{mid}{\textbf{9}} & \cellcolor{lower}{2} & \cellcolor{lower}{7} & \cellcolor{lower}{5} & \cellcolor{lower}{4}\\
\bottomrule
\end{tabular}}
\caption{Idioms}
\label{tab:idioms_layers}
\end{subtable}

\bigskip

\begin{subtable}{1\textwidth}
\sisetup{table-format=-1.2}   
\centering
\resizebox{\textwidth}{!}{%
\begin{tabular}{@{}l*{18}{c}@{}}
\toprule
\textbf{Task} & \multicolumn{3}{c}{\textbf{EN}} & \multicolumn{3}{c}{\textbf{DE}} & \multicolumn{3}{c}{\textbf{NL}} & \multicolumn{3}{c}{\textbf{PL}} & \multicolumn{3}{c}{\textbf{RU}} & \multicolumn{3}{c}{\textbf{UK}} \\
\cmidrule(lr){2-4}\cmidrule(lr){5-7}\cmidrule(lr){8-10}\cmidrule(lr){11-13}\cmidrule(lr){14-16}\cmidrule(lr){17-19}
 & T1 & T2 & T3 & T1 & T2 & T3 & T1 & T2 & T3 & T1 & T2 & T3 & T1 & T2 & T3 & T1 & T2 & T3 \\
\midrule
Pre-trained & -- & -- & -- & -- & -- & -- & -- & -- & -- & \cellcolor{lower}{3} & \cellcolor{lower}{2} & \cellcolor{lower}{1} & \cellcolor{lower}{2} & \cellcolor{lower}{4} & \cellcolor{lower}{3} & \cellcolor{lower}{3} & \cellcolor{lower}{2} & \cellcolor{mid}{\textbf{12}}\\
\midrule
DepRel & -- & -- & -- & -- & -- & -- & -- & -- & -- & \cellcolor{lower}{3} & \cellcolor{lower}{2} & \cellcolor{lower}{1} & \cellcolor{lower}{2} & \cellcolor{lower}{3} & \cellcolor{lower}{5} & \cellcolor{lower}{3} & \cellcolor{lower}{2} & \cellcolor{mid}{\textbf{12}}\\
POS & -- & -- & -- & -- & -- & -- & -- & -- & -- & \cellcolor{lower}{3} & \cellcolor{lower}{2} & \cellcolor{lower}{1} & \cellcolor{mid}{\textbf{11}} & \cellcolor{lower}{2} & \cellcolor{mid}{\textbf{12}} & \cellcolor{lower}{3} & \cellcolor{lower}{2} & \cellcolor{mid}{\textbf{12}}\\
\midrule
NER & -- & -- & -- & -- & -- & -- & -- & -- & -- & \cellcolor{lower}{3} & \cellcolor{lower}{2} & \cellcolor{lower}{1} & \cellcolor{lower}{2} & \cellcolor{lower}{3} & \cellcolor{lower}{6} & \cellcolor{lower}{3} & \cellcolor{lower}{2} & \cellcolor{lower}{1}\\
Topic & -- & -- & -- & -- & -- & -- & -- & -- & -- & \cellcolor{lower}{3} & \cellcolor{lower}{2} & \cellcolor{lower}{1} & \cellcolor{lower}{2} & \cellcolor{lower}{4} & \cellcolor{lower}{3} & \cellcolor{lower}{3} & \cellcolor{lower}{2} & \cellcolor{lower}{1}\\
\bottomrule
\end{tabular}}
\caption{Microsyntactic Units}
\label{tab:msu_layers}
\end{subtable}

\caption{Top three layers with highest attention percentage allocated to idioms and microsyntactic units (MSUs) across six languages. The layers with the highest attention percentages are labeled as T1 (highest), T2 (second highest), and T3 (third highest). Lower layers (1-8) are marked in blue color and middle layers (9-16) in yellow color and bold font to illustrate where the model focuses its attention.} \label{tab:top_layers}
\end{table*}

\subsection{Pre-trained vs. Fine-tuned Models}

\begin{figure*}
\centering\includegraphics[width=1\linewidth]{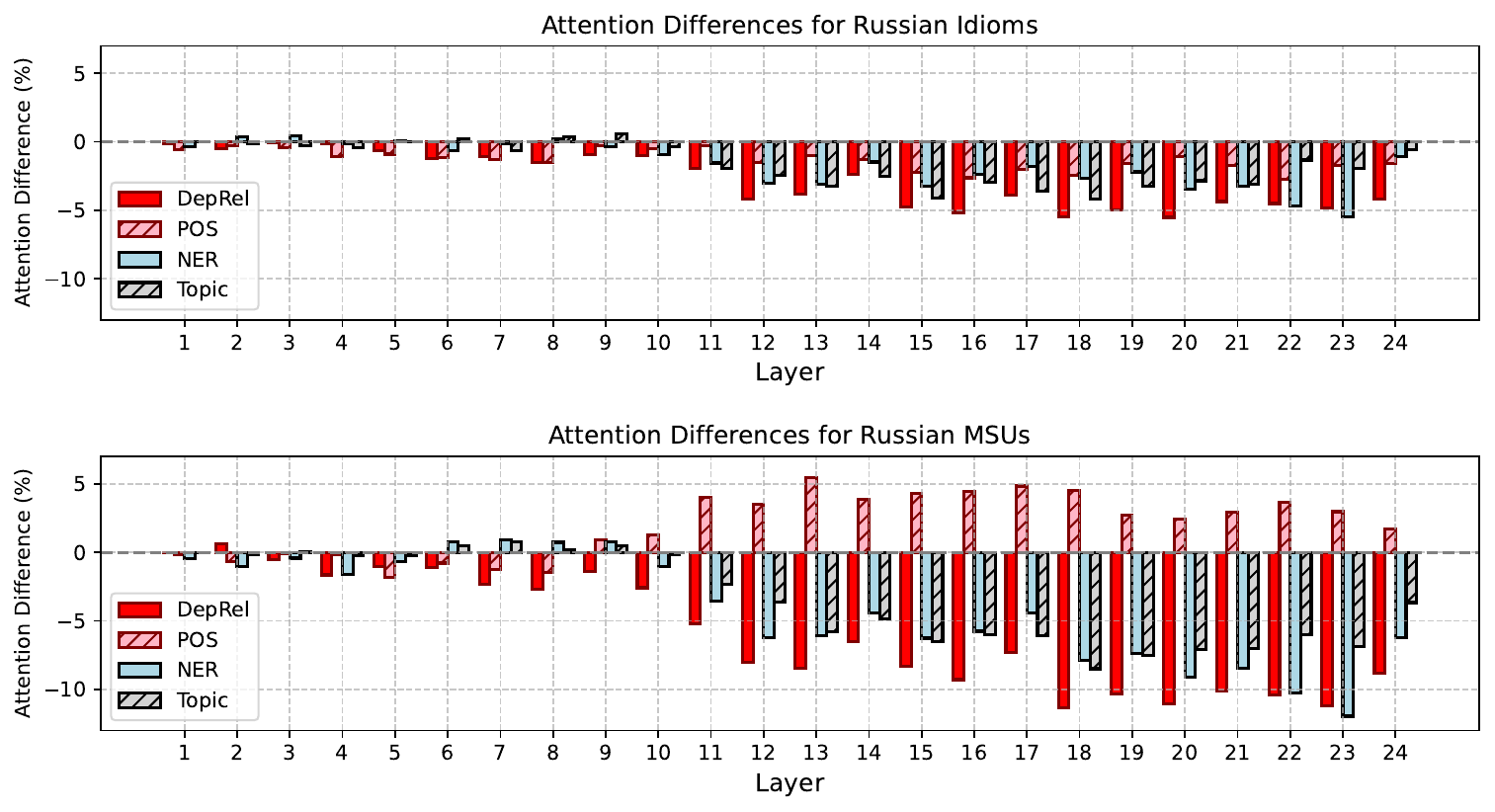}
     \caption{Differences in Attention Patterns for Idioms and Microsyntactic Units (MSUs) in Russian.
The bars show layer-wise changes in attention percentage after fine-tuning on syntactic tasks (Dependency Relation Classification – DepRel, Part-of-Speech Tagging – POS) and semantic tasks (Named Entity Recognition – NER, Topic Classification – Topic). Positive values indicate increased attention and negative values show decreased attention compared to the pre-trained model.} 
    \label{fig:attention_differences}
\end{figure*}

Figure \ref{fig:attention_results} shows the percentage of attention scores directed towards MWEs by layer in PL, UK, and EN. The figure compares the pre-trained models with models fine-tuned on syntactic and semantic tasks. As can be seen from the figure, in the pre-trained models, the attention directed towards both idioms and MSUs is more uniform across middle and upper layers, indicating a tendency towards general-purpose representation. The models fine-tuned on both syntactic and semantic tasks demonstrate noticeably higher attention peaks. This implies that fine-tuning improves the model's ability to concentrate on MWEs in general.

In addition to attention from other tokens in the sentence to MWEs, we analyzed the attention scores within the tokens of MWEs. Figure~\ref{fig:self_attention_plots} illustrates the percentage of attention scores within idioms and MSUs by layers in the DE, NL, and RU models. 
Here, we can again observe stronger attention peaks in the fine-tuned models as compared to the pre-trained models.

According to attention scores both within MWE and from context to MWE, models fine-tuned on semantic tasks -- NER and Topic -- show an increase in attention to MWEs in the higher layers compared to the pre-trained model. This is consistent across all six languages studied except for RU.

Figure \ref{fig:attention_differences} shows the differences in the percentage of average attention values from the pre-trained model across layers for two types of MWEs in RU. From the figure, we can see that fine-tuning leads to mostly decreased attention to both idioms and MSUs across most layers, and particularly in higher layers. Fine-tuning on syntactic tasks (Topic and Dep\-Rel) leads to a sharp decrease in middle to upper layers, except for POS task when processing MSUs. 
Fine-tuning on semantic tasks (NER and Topic) results in mixed changes. For both NER and Topic, attention to idioms and MSUs decreases in most layers but slightly increases in some lower to middle layers. 

These mixed changes suggest that fine-tuning on NER and Topic classification may cause the model to redistribute attention in a more nuanced way, possibly balancing between capturing named entity-specific information and broader contextual cues.

The consistent decrease in attention across most layers and tasks may indicate that fine-tuning reduces the model's sensitivity to MWEs. The changes in attention are especially pronounced in higher layers. Since higher layers are often associated with capturing abstract and semantic information, as mentioned in Section \ref{related_work}, their decreased attention suggests a possible trade-off between task performance and the model's ability to handle MWEs.

\subsection{Idioms vs. Microsyntactic Units}

Regardless of whether the model is only pre-trained or has been fine-tuned, a consistent pattern where attention peaks at the lower layers (3-4) when processing MSUs is observed across all models for all languages.

The presence of attention peaks in the lower layers for MWEs, and especially for MSUs in both figures, suggests that these units are more closely associated with syntactic processing. As stated before, lower layers of neural networks are typically more focused on syntactic features of a language since they process more structural aspects of the input data before passing higher-level semantic information to upper layers.

For MSUs, attention percentage is in general more varied across layers in models fine-tuned on syntactic tasks, while semantic tasks lead to a flatter, more uniform distribution.
In Figure \ref{fig:attention_differences}, we can see that attention to MSUs increases in middle and upper layers when fine-tuned on the POS task. The other syntactic task, DepRel, does not show the same pattern. In contrast, the attention scores for MSUs drop even more compared to idioms, except for a slight increase in layer 2. This could be because DepRel relies more on understanding grammatical relations rather than the specific syntactic categories like POS.

\subsection{Comparison between Language Groups}

With the Germanic languages (EN, DE, NL), the models produce more uniform attention patterns towards idioms within MWE and from context to MWE. This may be attributed to lower morphological complexity in Germanic languages, allowing the models to adapt more uniformly during fine-tuning.

In contrast, the models for Slavic languages (PL, RU, UK), especially for PL, display more varied attention patterns, as can be seen in Figure \ref{fig:attention_results}. PL also stands out from all other languages by the presence of a large attention peak in the lower layers, exactly at the layers where the attention drops for other languages. Since this is only observed when processing PL idioms and not PL MSUs, such anomaly is more likely to be related to the dataset rather than the language or the model.

Table \ref{tab:top_layers} provides an overview of top three layers with highest attention percentage towards idioms and MSUs in six languages. The table shows that, in general, for EN, NL, RU, and UK, the layers with highest attention to MWEs are spread across both lower and middle layers. For DE and PL, all MWEs receive high attention in the lower layers. We can see, however, that MSUs are mostly attended to in lower layers as opposed to idioms, which is expected given that idioms typically attract more attention in higher layers.

\section{Conclusion}

In this study, we analyzed the attention patterns of fine-tuned BERT-based models in relation to idioms and MSUs across six Indo-European languages from two language groups. By extending the methodology of \citet{jang-etal-2024-study} to MWEs, we demonstrated that fine-tuning on syntactic and semantic tasks significantly affects how models allocate attention to different types of MWEs.

Our results indicate that:

\begin{itemize}
    \item In general, models fine-tuned on syntactic tasks exhibit increased attention to MSUs in lower to middle layers, in accordance with syntactic processing requirements.
 \item  Models fine-tuned on semantic tasks show a tendency to distribute attention more evenly across layers, which could reflect a need for integrating information across different layers.
\item Cross-linguistic differences exist both between Germanic and Slavic languages, as well as across languages of the same language group. This underscores the complexity of how transformer models manage attention distribution. 

\item While there is a general trend towards decreased attention in syntactic tasks and more evenly distributed attention in semantic tasks, the anomalies highlight the non-uniform behavior of attention mechanisms in BERT-based models.
\end{itemize} 

These findings suggest that attention mechanisms in transformer models adapt during fine-tuning to better handle complex linguistic phenomena, according to the linguistic properties of the target language and task.

Future work could explore larger and more diverse datasets, different model architectures, and additional languages to build upon our findings and expand the understanding of how neural models process MWEs.

\subsection*{Limitations}

While our study provides valuable insights into how BERT-based models process MWEs across different languages and tasks, several limitations should be acknowledged.

Our research utilized idiom datasets obtained from various sources, which differ in composition and definition. The concept of an idiom itself lacks a precise, universally accepted definition, which might lead to inconsistencies in interpreting the results. This variability could affect the generalizability of our findings, as the models' performance might be influenced by the idiosyncrasies of the datasets used. Our dataset of microsyntactic units includes only the Slavic language groups, which could also bias the analysis.

Moreover, while our study focuses on semantic tasks like Named Entity Recognition and Topic Classification, which require semantic understanding, these tasks may involve only shallow semantic processing. This limitation could affect the extent to which fine-tuning on these tasks enhances the model's attention to idioms. The performance of BERT-based models on other semantic NLP tasks, such as summarization and paraphrasing, could provide additional insights into their ability to handle MWEs.

\section*{Acknowledgments}
We thank the anonymous reviewers for their constructive feedback. This research is funded by the Deutsche Forschungsgemeinschaft (DFG, German Research Foundation), Project-ID 232722074 – SFB 1102.
\bibliography{custom}

\begin{thebibliography}{38}
\providecommand{\natexlab}[1]{#1}

\bibitem[{rus(2003--2023)}]{ruscorpora}
 2003--2023.
\newblock Russian {N}ational {C}orpus.
\newblock Accessed 25.09.2024.

\bibitem[{Adelani et~al.(2023)Adelani, Liu, Shen, Vassilyev, Alabi, Mao, Gao, and Lee}]{adelani2023sib200}
David~Ifeoluwa Adelani, Hannah Liu, Xiaoyu Shen, Nikita Vassilyev, Jesujoba~O. Alabi, Yanke Mao, Haonan Gao, and Annie En-Shiun Lee. 2023.
\newblock \href {https://arxiv.org/abs/2309.07445} {Sib-200: A simple, inclusive, and big evaluation dataset for topic classification in 200+ languages and dialects}.
\newblock \emph{Preprint}, arXiv:2309.07445.

\bibitem[{Aharodnik et~al.(2018)Aharodnik, Feldman, and Peng}]{Aharodnik2018DesigningAR}
Katsiaryna Aharodnik, Anna Feldman, and Jing Peng. 2018.
\newblock \href {https://api.semanticscholar.org/CorpusID:21722705} {Designing a russian idiom-annotated corpus}.
\newblock In \emph{International Conference on Language Resources and Evaluation}.

\bibitem[{Avgustinova and Iomdin(2019)}]{Avgustinova2019}
Tania Avgustinova and Leonid Iomdin. 2019.
\newblock \href {https://doi.org/10.1007/978-3-030-30135-4_2} {\emph{Towards a Typology of Microsyntactic Constructions}}, volume 11755 of Lecture Notes in Computer Science. Springer, Cham., pages 15--30.

\bibitem[{Avram et~al.(2023)Avram, Barbu~Mititelu, and Cercel}]{avram-etal-2023-romanian}
Andrei Avram, Verginica Barbu~Mititelu, and Dumitru-Clementin Cercel. 2023.
\newblock \href {https://doi.org/10.18653/v1/2023.mwe-1.4} {{R}omanian multiword expression detection using multilingual adversarial training and lateral inhibition}.
\newblock In \emph{Proceedings of the 19th Workshop on Multiword Expressions (MWE 2023)}, pages 7--13, Dubrovnik, Croatia. Association for Computational Linguistics.

\bibitem[{Baldwin and Kim(2010)}]{baldwin2010multiword}
Timothy Baldwin and Su~Nam Kim. 2010.
\newblock Multiword expressions.
\newblock In Nitin Indurkhya and Fred~J. Damerau, editors, \emph{Handbook of Natural Language Processing}, pages 267--292. Chapman and Hall/CRC.

\bibitem[{Baranov and Dobrovolsky(2015)}]{baranov2015}
A~Baranov and D~Dobrovolsky, editors. 2015.
\newblock \emph{Academic Dictionary of Russian Phraseology}, 2 edition.
\newblock LEKSRUS, Moscow.

\bibitem[{Boisson et~al.(2022)Boisson, Camacho-Collados, and Espinosa-Anke}]{boisson-etal-2022-cardiffnlp}
Joanne Boisson, Jose Camacho-Collados, and Luis Espinosa-Anke. 2022.
\newblock \href {https://doi.org/10.18653/v1/2022.semeval-1.20} {{C}ardiff{NLP}-metaphor at {S}em{E}val-2022 task 2: Targeted fine-tuning of transformer-based language models for idiomaticity detection}.
\newblock In \emph{Proceedings of the 16th International Workshop on Semantic Evaluation (SemEval-2022)}, pages 169--177, Seattle, United States. Association for Computational Linguistics.

\bibitem[{Bui and Savary(2024)}]{bui-savary-2024-cross}
Van-Tuan Bui and Agata Savary. 2024.
\newblock \href {https://aclanthology.org/2024.lrec-main.374} {Cross-type {F}rench multiword expression identification with pre-trained masked language models}.
\newblock In \emph{Proceedings of the 2024 Joint International Conference on Computational Linguistics, Language Resources and Evaluation (LREC-COLING 2024)}, pages 4198--4204, Torino, Italia. ELRA and ICCL.

\bibitem[{Chan et~al.(2020)Chan, Schweter, and M{\"o}ller}]{chan-etal-2020-germans}
Branden Chan, Stefan Schweter, and Timo M{\"o}ller. 2020.
\newblock \href {https://doi.org/10.18653/v1/2020.coling-main.598} {{G}erman{'}s next language model}.
\newblock In \emph{Proceedings of the 28th International Conference on Computational Linguistics}, pages 6788--6796, Barcelona, Spain (Online). International Committee on Computational Linguistics.

\bibitem[{Dankers et~al.(2022)Dankers, Lucas, and Titov}]{dankers-etal-2022-transformer}
Verna Dankers, Christopher Lucas, and Ivan Titov. 2022.
\newblock \href {https://doi.org/10.18653/v1/2022.acl-long.252} {Can transformer be too compositional? analysing idiom processing in neural machine translation}.
\newblock In \emph{Proceedings of the 60th Annual Meeting of the Association for Computational Linguistics (Volume 1: Long Papers)}, pages 3608--3626, Dublin, Ireland. Association for Computational Linguistics.

\bibitem[{Delobelle and Remy(2023)}]{delobelle2023robbert2023conversion}
P~Delobelle and F~Remy. 2023.
\newblock \href {https://clin33.uantwerpen.be/abstract/robbert-2023-keeping-dutch-language-models-up-to-date-at-a-lower-cost-thanks-to-model-conversion/} {Robbert-2023: Keeping dutch language models up-to-date at a lower cost thanks to model conversion}.

\bibitem[{Devlin et~al.(2019)Devlin, Chang, Lee, and Toutanova}]{devlin-etal-2019-bert}
Jacob Devlin, Ming-Wei Chang, Kenton Lee, and Kristina Toutanova. 2019.
\newblock \href {https://doi.org/10.18653/v1/N19-1423} {{BERT}: Pre-training of deep bidirectional transformers for language understanding}.
\newblock In \emph{Proceedings of the 2019 Conference of the North {A}merican Chapter of the Association for Computational Linguistics: Human Language Technologies, Volume 1 (Long and Short Papers)}, pages 4171--4186, Minneapolis, Minnesota. Association for Computational Linguistics.

\bibitem[{Fadaee et~al.(2018)Fadaee, Bisazza, and Monz}]{fadaee2018examining}
Marzieh Fadaee, Arianna Bisazza, and Christof Monz. 2018.
\newblock Examining the tip of the iceberg: A data set for idiom translation.
\newblock In \emph{Proceedings of the Eleventh International Conference on Language Resources and Evaluation ({LREC} 2018)}, Miyazaki, Japan. European Language Resources Association (ELRA).

\bibitem[{Haltiuk and Smywi{\'n}ski-Pohl(2024)}]{haltiuk-smywinski-pohl-2024-liberta}
Mykola Haltiuk and Aleksander Smywi{\'n}ski-Pohl. 2024.
\newblock \href {https://aclanthology.org/2024.unlp-1.14} {{L}i{BERT}a: Advancing {U}krainian language modeling through pre-training from scratch}.
\newblock In \emph{Proceedings of the Third Ukrainian Natural Language Processing Workshop (UNLP) @ LREC-COLING 2024}, pages 120--128, Torino, Italia. ELRA and ICCL.

\bibitem[{Iomdin(2015)}]{iomdin2015}
Leonid Iomdin. 2015.
\newblock Microsyntactic constructions formed by the {Russian} word raz.
\newblock \emph{SLAVIA cˇasopis pro slovanskou filologii}, 84(3).

\bibitem[{Iomdin(2016)}]{iomdin-2016-microsyntactic}
Leonid Iomdin. 2016.
\newblock \href {https://aclanthology.org/W16-3803} {Microsyntactic phenomena as a computational linguistics issue}.
\newblock In \emph{Proceedings of the Workshop on Grammar and Lexicon: interactions and interfaces ({G}ram{L}ex)}, pages 8--17, Osaka, Japan. The COLING 2016 Organizing Committee.

\bibitem[{Jang et~al.(2024)Jang, Byun, and Shin}]{jang-etal-2024-study}
Dongjun Jang, Sungjoo Byun, and Hyopil Shin. 2024.
\newblock \href {https://aclanthology.org/2024.lrec-main.148} {A study on how attention scores in the {BERT} model are aware of lexical categories in syntactic and semantic tasks on the {GLUE} benchmark}.
\newblock In \emph{Proceedings of the 2024 Joint International Conference on Computational Linguistics, Language Resources and Evaluation (LREC-COLING 2024)}, pages 1684--1689, Torino, Italia. ELRA and ICCL.

\bibitem[{Kurfal{\i}(2020)}]{kurfali-2020-travis}
Murathan Kurfal{\i}. 2020.
\newblock \href {https://aclanthology.org/2020.mwe-1.18} {{TRAVIS} at {PARSEME} shared task 2020: How good is (m){BERT} at seeing the unseen?}
\newblock In \emph{Proceedings of the Joint Workshop on Multiword Expressions and Electronic Lexicons}, pages 136--141, online. Association for Computational Linguistics.

\bibitem[{Machálek(2020)}]{machalek2020kontext}
Tomáš Machálek. 2020.
\newblock Kontext: Advanced and flexible corpus query interface.
\newblock In \emph{Proceedings of the Twelfth Language Resources and Evaluation Conference}, pages 7003--7008, Marseille, France. European Language Resources Association.

\bibitem[{Masini(2019)}]{MultiWordExpressionsandMorphology}
Francesca Masini. 2019.
\newblock \href {https://doi.org/10.1093/acrefore/9780199384655.013.611} {Multi-word expressions and morphology}.
\newblock Oxford University Press, Oxford.

\bibitem[{Mileti{\'c} and Walde(2024)}]{miletic-walde-2024-semantics}
Filip Mileti{\'c} and Sabine Schulte~im Walde. 2024.
\newblock \href {https://doi.org/10.1162/tacl_a_00657} {Semantics of multiword expressions in transformer-based models: A survey}.
\newblock \emph{Transactions of the Association for Computational Linguistics}, 12:593--612.

\bibitem[{Mroczkowski et~al.(2021)Mroczkowski, Rybak, Wr{\'o}blewska, and Gawlik}]{mroczkowski-etal-2021-herbert}
Robert Mroczkowski, Piotr Rybak, Alina Wr{\'o}blewska, and Ireneusz Gawlik. 2021.
\newblock \href {https://www.aclweb.org/anthology/2021.bsnlp-1.1} {{H}er{BERT}: Efficiently pretrained transformer-based language model for {P}olish}.
\newblock In \emph{Proceedings of the 8th Workshop on Balto-Slavic Natural Language Processing}, pages 1--10, Kyiv, Ukraine. Association for Computational Linguistics.

\bibitem[{Nivre et~al.(2020)Nivre, de~Marneffe, Ginter, Haji{\v{c}}, Manning, Pyysalo, Schuster, Tyers, and Zeman}]{nivre-etal-2020-universal}
Joakim Nivre, Marie-Catherine de~Marneffe, Filip Ginter, Jan Haji{\v{c}}, Christopher~D. Manning, Sampo Pyysalo, Sebastian Schuster, Francis Tyers, and Daniel Zeman. 2020.
\newblock \href {https://aclanthology.org/2020.lrec-1.497} {{U}niversal {D}ependencies v2: An evergrowing multilingual treebank collection}.
\newblock In \emph{Proceedings of the Twelfth Language Resources and Evaluation Conference}, pages 4034--4043, Marseille, France. European Language Resources Association.

\bibitem[{Pan et~al.(2017)Pan, Zhang, May, Nothman, Knight, and Ji}]{pan-etal-2017-cross}
Xiaoman Pan, Boliang Zhang, Jonathan May, Joel Nothman, Kevin Knight, and Heng Ji. 2017.
\newblock \href {https://doi.org/10.18653/v1/P17-1178} {Cross-lingual name tagging and linking for 282 languages}.
\newblock In \emph{Proceedings of the 55th Annual Meeting of the Association for Computational Linguistics (Volume 1: Long Papers)}, pages 1946--1958, Vancouver, Canada. Association for Computational Linguistics.

\bibitem[{Rambelli et~al.(2023)Rambelli, Chersoni, Senaldi, Blache, and Lenci}]{rambelli-etal-2023-frequent}
Giulia Rambelli, Emmanuele Chersoni, Marco S.~G. Senaldi, Philippe Blache, and Alessandro Lenci. 2023.
\newblock \href {https://doi.org/10.18653/v1/2023.mwe-1.13} {Are frequent phrases directly retrieved like idioms? an investigation with self-paced reading and language models}.
\newblock In \emph{Proceedings of the 19th Workshop on Multiword Expressions (MWE 2023)}, pages 87--98, Dubrovnik, Croatia. Association for Computational Linguistics.

\bibitem[{Shwartz and Dagan(2019)}]{shwartz-dagan-2019-still}
Vered Shwartz and Ido Dagan. 2019.
\newblock \href {https://doi.org/10.1162/tacl_a_00277} {Still a pain in the neck: Evaluating text representations on lexical composition}.
\newblock \emph{Transactions of the Association for Computational Linguistics}, 7:403--419.

\bibitem[{Tan and Jiang(2021)}]{tan-jiang-2021-bert}
Minghuan Tan and Jing Jiang. 2021.
\newblock \href {https://aclanthology.org/2021.ranlp-1.156/} {Does {BERT} understand idioms? a probing-based empirical study of {BERT} encodings of idioms}.
\newblock In \emph{Proceedings of the International Conference on Recent Advances in Natural Language Processing (RANLP 2021)}, pages 1397--1407, Held Online. INCOMA Ltd.

\bibitem[{Tedeschi et~al.(2022)Tedeschi, Martelli, and Navigli}]{tedeschi-etal-2022-id10m}
Simone Tedeschi, Federico Martelli, and Roberto Navigli. 2022.
\newblock \href {https://doi.org/10.18653/v1/2022.findings-naacl.208} {{ID}10{M}: Idiom identification in 10 languages}.
\newblock In \emph{Findings of the Association for Computational Linguistics: NAACL 2022}, pages 2715--2726, Seattle, United States. Association for Computational Linguistics.

\bibitem[{Tenney et~al.(2019)Tenney, Das, and Pavlick}]{tenney2019bertrediscoversclassicalnlp}
Ian Tenney, Dipanjan Das, and Ellie Pavlick. 2019.
\newblock \href {https://arxiv.org/abs/1905.05950} {Bert rediscovers the classical nlp pipeline}.
\newblock \emph{Preprint}, arXiv:1905.05950.

\bibitem[{Tian et~al.(2023)Tian, James, and Son}]{tian-etal-2023-idioms}
Ye~Tian, Isobel James, and Hye Son. 2023.
\newblock \href {https://doi.org/10.18653/v1/2023.starsem-1.16} {How are idioms processed inside transformer language models?}
\newblock In \emph{Proceedings of the 12th Joint Conference on Lexical and Computational Semantics (*SEM 2023)}, pages 174--179, Toronto, Canada. Association for Computational Linguistics.

\bibitem[{Walsh et~al.(2022)Walsh, Lynn, and Foster}]{walsh-etal-2022-berts}
Abigail Walsh, Teresa Lynn, and Jennifer Foster. 2022.
\newblock \href {https://aclanthology.org/2022.mwe-1.13} {A {BERT}{'}s eye view: Identification of {I}rish multiword expressions using pre-trained language models}.
\newblock In \emph{Proceedings of the 18th Workshop on Multiword Expressions @LREC2022}, pages 89--99, Marseille, France. European Language Resources Association.

\bibitem[{Wang et~al.(2018)Wang, Singh, Michael, Hill, Levy, and Bowman}]{wang-etal-2018-glue}
Alex Wang, Amanpreet Singh, Julian Michael, Felix Hill, Omer Levy, and Samuel Bowman. 2018.
\newblock \href {https://doi.org/10.18653/v1/W18-5446} {{GLUE}: A multi-task benchmark and analysis platform for natural language understanding}.
\newblock In \emph{Proceedings of the 2018 {EMNLP} Workshop {B}lackbox{NLP}: Analyzing and Interpreting Neural Networks for {NLP}}, pages 353--355, Brussels, Belgium. Association for Computational Linguistics.

\bibitem[{Warren(2005)}]{warren_2005}
Beatrice Warren. 2005.
\newblock A model of idiomaticity.
\newblock \emph{Nordic Journal of English Studies}, 4:35--54.

\bibitem[{Wolf et~al.(2020)Wolf, Debut, Sanh, Chaumond, Delangue, Moi, Cistac, Rault, Louf, Funtowicz, Davison, Shleifer, von Platen, Ma, Jernite, Plu, Xu, Scao, Gugger, Drame, Lhoest, and Rush}]{wolf-etal-2020-transformers}
Thomas Wolf, Lysandre Debut, Victor Sanh, Julien Chaumond, Clement Delangue, Anthony Moi, Pierric Cistac, Tim Rault, Rémi Louf, Morgan Funtowicz, Joe Davison, Sam Shleifer, Patrick von Platen, Clara Ma, Yacine Jernite, Julien Plu, Canwen Xu, Teven~Le Scao, Sylvain Gugger, Mariama Drame, Quentin Lhoest, and Alexander~M. Rush. 2020.
\newblock \href {https://www.aclweb.org/anthology/2020.emnlp-demos.6} {Transformers: State-of-the-art natural language processing}.
\newblock In \emph{Proceedings of the 2020 Conference on Empirical Methods in Natural Language Processing: System Demonstrations}, pages 38--45, Online. Association for Computational Linguistics.

\bibitem[{Zaitova et~al.(2023)Zaitova, Stenger, and Avgustinova}]{zaitova-etal-2023-microsyntactic}
Iuliia Zaitova, Irina Stenger, and Tania Avgustinova. 2023.
\newblock \href {https://aclanthology.org/2023.ranlp-1.134} {Microsyntactic unit detection using word embedding models: Experiments on {S}lavic languages}.
\newblock In \emph{Proceedings of the 14th International Conference on Recent Advances in Natural Language Processing}, pages 1265--1273, Varna, Bulgaria. INCOMA Ltd., Shoumen, Bulgaria.

\bibitem[{Zeng and Bhat(2021)}]{10.1162/tacl_a_00442}
Ziheng Zeng and Suma Bhat. 2021.
\newblock \href {https://doi.org/10.1162/tacl_a_00442} {{Idiomatic Expression Identification using Semantic Compatibility}}.
\newblock \emph{Transactions of the Association for Computational Linguistics}, 9:1546--1562.

\bibitem[{Zmitrovich et~al.(2023)Zmitrovich, Abramov, Kalmykov, Tikhonova, Taktasheva, Astafurov, Baushenko, Snegirev, Shavrina, Markov, Mikhailov, and Fenogenova}]{zmitrovich2023family}
Dmitry Zmitrovich, Alexander Abramov, Andrey Kalmykov, Maria Tikhonova, Ekaterina Taktasheva, Danil Astafurov, Mark Baushenko, Artem Snegirev, Tatiana Shavrina, Sergey Markov, Vladislav Mikhailov, and Alena Fenogenova. 2023.
\newblock \href {https://arxiv.org/abs/2309.10931} {A family of pretrained transformer language models for russian}.
\newblock \emph{Preprint}, arXiv:2309.10931.

\end{thebibliography}



\end{document}